\documentclass[runningheads]{llncs}
\usepackage[T1]{fontenc}
\usepackage{xcolor}
\definecolor{miccai}{rgb}{0.21,0.49,0.74}
\usepackage[pagebackref,breaklinks,colorlinks,allcolors=miccai]{hyperref}
\usepackage{placeins}
\usepackage{multirow}
\usepackage{bm}
\usepackage{float}
\usepackage{hyperref} 
\usepackage{booktabs}
\usepackage{enumitem}
\usepackage{amssymb}
\usepackage{xspace}
\usepackage{xurl}

\usepackage[most]{tcolorbox}
\tcbset{colback=gray!3,colframe=gray!70!black,boxrule=0.4pt,arc=1.2pt,left=4pt,right=4pt,top=4pt,bottom=4pt}
\newtcblisting{promptbox}[1]{title=#1,listing only,breakable,fonttitle=\bfseries\footnotesize,listing options={basicstyle=\ttfamily\footnotesize,breaklines=true, breakatwhitespace=true, breakautoindent=false, breakindent=0pt, columns=fixed}}
\usepackage{marvosym}
\usepackage{graphicx,verbatim}
\begin{document}

\title{FRAC-MAS: A Safe and Explainable Multi-Agent System for Fracture Diagnosis}
\titlerunning{FRAC-MAS}
%


\author{
Hardik Iyer\inst{1}\textsuperscript{*} \and
Tirath Bhathawala\inst{2} \and
Mihir Panchal\inst{3} \and
Ying-Jung Chen\inst{4} \and
Kiran Bhowmick\inst{1} \and
Pankaj Sonawane\inst{1} \and
Meera Narvekar\inst{1}
}


\authorrunning{Iyer et al.}
 
\institute{
Dwarkadas J Sanghvi College of Engineering, Mumbai, India\\
\and
University of Amsterdam, Amsterdam, Netherlands\\
\and
National University of Singapore, Singapore\\
\and
Georgia Institute of Technology, Atlanta, USA\\
\email{hardikiyer17@gmail.com},
\email{tirath.bhathawala@student.uva.nl},
\email{mihir@comp.nus.edu.sg},
\email{yingjungcd@gmail.com},
\email{kiran.bhowmick@djsce.ac.in, pankaj.sonawane@djsce.ac.in, meera.narvekar@djsce.ac.in}
}
 
  
\maketitle              
\begin{abstract}

Fracture detection and its clinical interpretability see notable improvements when deep vision models are integrated with agentic AI architectures. While deep learning models achieve high diagnostic performance, their black-box nature limits clinical adoption. We propose FRAC-MAS, an agentic AI system for automated, explainable, and safe bone fracture detection. The framework combines a stacked ensemble of four vision models with conformal prediction to produce statistically grounded differential diagnoses, while a multi-agent workflow performs independent verification, retrieves clinical guidelines, and generates patient-friendly reports. A pipeline-depth ablation study confirms that our multi-agent critic triages 86.6\% of cases into a high-confidence auto-confirmed cohort while escalating uncertain cases, outperforming a single-agent baseline. Patient preference studies against Llama, MedGemma, and Gemini further demonstrate significantly more comprehensible clinical reports. These results suggest that integrating multi-agent critics with conformal guarantees enables safer radiology triage while preserving clinician oversight. More broadly, FRAC-MAS demonstrates how cooperative agentic architectures can serve as auditable, human-in-the-loop decision support systems for safety-critical healthcare. Our code is available at \url{https://github.com/hardik1712/FRAC-MAS}, and the website is available at \url{https://frac-mas.vercel.app}.


\end{abstract}
\section{Introduction}\label{sec:intro} 

Musculoskeletal injuries, especially bone fractures, represent a substantial proportion of cases in emergency and orthopedic departments~\cite{yan2025global}. Interpreting plain radiographs is essential for accurate fracture diagnosis and management, yet it remains challenging due to subtle fracture morphologies, inter-observer variability, and the high volume of cases seen in clinical practice~\cite{pinto2018traumatic}. Undetected fractures frequently progress to malunion, avascular necrosis, and permanent disability, while missed diagnoses drive avoidable surgeries, prolonged hospitalizations, and significant economic losses~\cite{flores2024economic}.

Studies report that about 2–3\% of all orthopedic injuries are missed at first presentation in trauma and emergency settings, and plain radiographs may not indicate the presence of fractures in about 10–20\% of some skull and skeletal injuries~\cite{gupta2024patterns}. Therefore, a non-trivial proportion of fractures remains under-detected at first contact~\cite{lindsey2018deep}. Explainable fracture-detection systems are increasingly being used to highlight the specific image regions that drive a fracture prediction~\cite{van2022explainable}. This allows clinicians to verify whether the model is focusing on plausible features such as cortical discontinuity or trabecular collapse rather than irrelevant artifacts, improving trust and supporting safer adoption of AI-assisted diagnosis~\cite{kassem2023explainable}.

Addressing these gaps, we propose FRAC-MAS, a multi-agent AI system for automated bone fracture detection with clinical explainability. The system emphasizes human-AI interaction by validating outputs with practicing orthopedic surgeons, since human oversight is critical in clinical AI pipelines to mitigate automation bias, ensure alignment with clinical values, and improve decision accuracy where AI alone cannot reliably handle ethical ambiguity or edge cases~\cite{olawade2026human}. Our multi-agent architecture bridges the gap between black-box deep learning predictions and clinical trust by decomposing the diagnostic pipeline into specialized, interpretable components, each independently verifiable and collectively designed for real-world deployment. 

This shift toward modular, specialized AI pipelines aligns with recent literature investigating targeted deep learning applications for fracture detection. Joonho Oh et al.~\cite{oh2023enhancing} demonstrated potential leaps in streamlining training processes and augmenting diagnostic precision in fracture detection, particularly focusing on X-ray images of the wrist bone using attention modules~\cite{woo2018cbam}. RAD-DINO has been established as a strong self-supervised vision backbone for medical imaging and has been systematically benchmarked across multiple radiology tasks, including musculoskeletal X-ray classification~\cite{perezgarcia2025raddino}. Furthermore, hybrid CNN-transformer and multi-stream architectures are necessary to capture both fine-grained local features and global contextual structure across heterogeneous radiographic views~\cite{hassan2025fracture,uddin2024mobilevit}. However, validating language model outputs generated from these visual features remains an open challenge; studies like Chung et al.~\cite{chung2025verifying} highlight the need for verification and LLM-as-a-judge protocols to validate documentation generated by clinical LLMs. Agentic extensions of these pipelines to real-world clinical deployment further validate retrieval-grounded, orchestrated architectures for trustworthy diagnostic AI~\cite{oettl2025artificial}.

Post-hoc interpretability methods are now integral to clinically deployable medical imaging AI, with Grad-CAM serving as the dominant visualization paradigm~\cite{selvaraju2017gradcam}.  Successfully applied to pelvic fracture detection~\cite{kassem2023explainable}, these developments establish that spatial explainability is essential for regulatory compliance and building clinician trust. Multi-agent systems (MAS) have demonstrated measurable superiority over single-agent baselines across clinical benchmarks. For example, Ying-Jung Chen et al.~\cite{chen2025enhancing} showed MAS achieves 59\% mortality prediction accuracy versus 56\% for single-agent systems, with mean length-of-stay error reduced from 5.82 to 4.37 days~\cite{chen2025enhancing,kim2024mdagents}. These results collectively motivate agent-orchestrated architectures wherein specialized roles including retrieval, reasoning, and critique are decomposed across coordinated agents to achieve diagnostic performance unattainable by monolithic models.

Briefly, the major contributions of this study are as follows. First, we design a multi-agent clinical decision support workflow, where a Patient Interaction Agent orchestrates Knowledge, Critic, and Educational agents to ground evidence, perform blind verification with human escalation, and generate patient-facing explanations with Grad-CAM overlays. Second, we introduce an ensemble with conformal prediction that resolves commonly confused fracture categories via stacking and converts outputs into differential diagnoses with explicit empirical coverage safety. Finally, we rigorously validate the agent layer via a novel pipeline-depth ablation study, proving that our multi-agent architecture successfully triages uncertain cases, and establish its patient-communication superiority via a preference survey against leading LLMs (Llama, MedGemma, Gemini).

\begin{figure*}[htb]
  \centering
  \includegraphics[width=\linewidth]{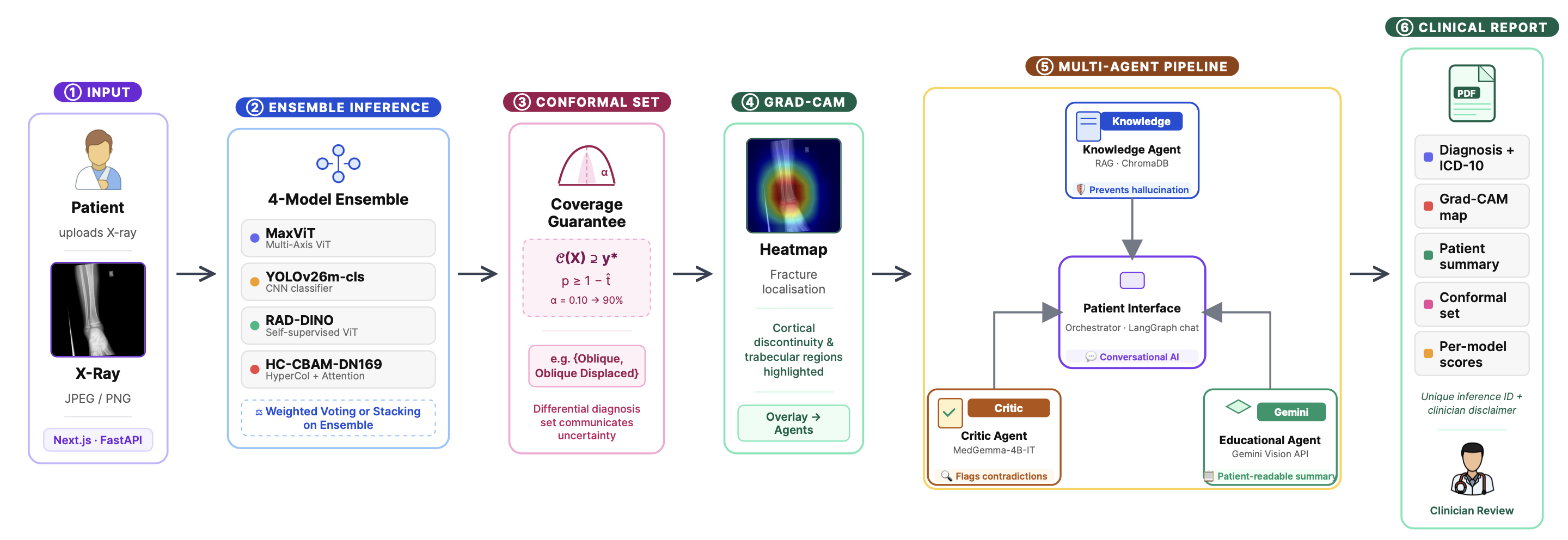}
  \caption{End-to-end system architecture: integrating ensemble inference, Grad-CAM localization, multi-agent reasoning, and conformal prediction for verifiable clinical report generation.}
  \label{fig:system_architecture}
\end{figure*} 

\vspace{-0.7cm}

\section{Methodology}\label{sec:methodology} 

\subsection{Dataset and Models Ensemble}\label{subsec:ensemble} 
The system is trained on an augmented version of the HBFMID dataset~\cite{parvin2024hbfmid}. We utilized this dataset because it categorizes radiographs by precise fracture morphology rather than mere anatomical location. We balanced and augmented the training set to 1,952 images across 8 categories, while maintaining unaugmented validation (106 images) and test (112 images) splits.

Our system employs an ensemble of 4 models to capture diverse fracture appearances: MaxViT~\cite{tu2022maxvit}, YOLOv26m-cls~\cite{sapkota2025yolo26}, RAD-DINO, and a custom HyperColumn-CBAM DenseNet-169~\cite{hariharan2015hypercolumns,huang2017densenet}. This combines multi-axis attention, self-supervised radiological domain features, and low-level edge detail preservation. Predictions are combined through a two-pass weighted soft voting scheme or a stacking meta-learner (logistic regression) to optimally disambiguate morphologically similar categories like Oblique and Transverse fractures.

\vspace{-0.4cm}

\subsection{Conformal Prediction}\label{subsec:conformal} 
To complement the ensemble output with a rigorous measure of uncertainty, the system wraps the ensemble output in a split conformal prediction set with a distribution-free empirical coverage guarantee~\cite{angelopoulos2023conformal}, calibrated on the validation set. Given a miscoverage level $\alpha$ (e.g., $\alpha = 0.10$ for 90\% coverage), every class whose probability exceeds a calibrated threshold enters the prediction set. This maps naturally to a differential diagnosis, communicating empirical uncertainty directly to downstream agents.

\vspace{-0.4cm}

\subsection{Multi-Agent System}\label{subsec:mulit-agent-system}

To make the prediction pipeline auditable end-to-end, we route every case through a four-agent workflow (see Fig. \ref{fig:system_architecture}). By decomposing the diagnostic process into specialized roles, the system prevents monolithic LLMs from hallucinating medical advice and forces all outputs to be explicitly grounded and independently verified.

\subsubsection{Patient Interface Agent.} 
The Patient Interface Agent acts as the entry point and primary orchestrator. It manages the conversational memory and evaluates the patient's context (e.g., age, reported pain levels). It utilizes a deterministic LangGraph state machine to route interactions: simple administrative queries are handled directly, while clinical assessments of uploaded radiographs invoke the downstream pipeline. This architecture ensures that diagnostic responses are generated strictly through the verified pipeline rather than through unbounded, parametric LLM generation. 

\subsubsection{Knowledge Agent.} 
The Knowledge Agent is responsible for anchoring the narrative to verified clinical sources, acting as an advanced Retrieval-Augmented Generation (RAG) module. It queries a locally hosted, curated ChromaDB vector store that contains structured clinical records for all 8 fracture classes. These records include standardized ICD-10 codes, expected recovery timelines, severity ratings, and official treatment guidelines from established orthopedic bodies (e.g., AO/OTA fracture classifications). For each radiograph case, the agent uses sentence-transformer embeddings to perform a dense semantic search. The retrieved top-k guidelines are then supplied as the absolute, non-negotiable context to a fast language model (\textit{gemini-2.5-flash-lite}~\cite{comanici2025gemini}), which synthesizes a focused clinical background that strictly avoids hallucinating non-retrieved treatments.

\subsubsection{Critic Agent.} 
Operating as the core safety mechanism, the Critic Agent prevents automation bias by performing a blind-first triage. Upon receiving an image, it produces an independent zero-shot diagnostic assessment \textit{before} seeing the ensemble's provisional label, preventing anchoring bias. It then evaluates the ensemble's output and issues a verdict $\in$ \{`yes`, `no`, `uncertain`\}. A rigorous hybrid consensus rule forces human escalation under three conditions: (1) if the Critic explicitly rejects the ensemble's diagnosis, (2) if the Critic's independent top prediction differs from the ensemble with a confidence greater than a preset safety threshold (default 0.6), or (3) if the conformal prediction set contains multiple classes, indicating intrinsic model ambiguity. This ensures the clinician remains the final arbiter for all difficult cases.

\subsubsection{Educational Agent.} 
Finally, the Educational Agent is tasked with patient communication. It receives the clinically verified diagnosis, the retrieved context from the Knowledge Agent, and the Grad-CAM~\cite{selvaraju2017gradcam} heatmap overlays generated by the vision ensemble. Using this rich context, it prompts a highly capable reasoning model (\textit{gemini-2.5-pro}~\cite{team2023gemini}) to generate a lay-friendly summary. The agent explicitly highlights the anatomical regions activated in the Grad-CAM heatmap to explain \textit{why} the diagnosis was made, translating complex radiographic findings into accessible language while providing clear next-steps guidance (e.g., immediate emergency room visits for displaced fractures versus scheduled orthopedic follow-ups for non-displaced injuries).

\section{Results and Analysis}\label{sec:results} 

\subsection{Pipeline Agent Ablation}\label{sec:pipeline_ablation} 

To quantify the marginal contribution of each architectural layer, we evaluated five incrementally richer arms of FRAC-MAS on the held-out test set (seed~42, $\alpha{=}0.10$, stacking meta-learner enabled). As shown in Table~\ref{tab:pipeline_ablation}, accuracy is invariant across pipeline depth because downstream components do not alter the raw argmax prediction. The single-agent baseline escalates every test case due to raw confidence thresholds, providing no practical automation benefit. In contrast, the multi-agent Critic selectively auto-confirms a large majority of cases, reducing unnecessary escalations significantly. The confirmed cohort achieves a strong safety margin over the raw ensemble baseline, demonstrating that the Critic successfully filters erroneous predictions. Finally, the full pipeline achieves a complete task-completion rate for verifiable, retrieval-grounded generation.

\begin{table}[htb]
\centering
\caption{Pipeline-depth ablation on the 112-sample test set ($\alpha{=}0.10$, stacking). \emph{Conf. Acc.} is accuracy on cases not flagged for human escalation. \textsuperscript{\textdagger}The Critic uses local MedGemma-4B-IT.}
\label{tab:pipeline_ablation}
\setlength{\tabcolsep}{4pt}
\resizebox{\columnwidth}{!}{%
\begin{tabular}{lcccccc}
\toprule
\textbf{Configuration} & \textbf{Acc.} (\%) & \textbf{Conf. Acc.} (\%) & \textbf{Escalation Rate} (\%) & \textbf{Set Size} & \textbf{Coverage} (\%) & \textbf{Know/Educ} (\%) \\
\midrule
Single-agent baseline & 91.1 & --- & 100.0 & --- & --- & --- \\
Ensemble only & 91.1 & --- & \phantom{1}0.0 & --- & --- & --- \\
+Conformal ($\alpha{=}0.10$) & 91.1 & --- & \phantom{1}0.0 & 1.08 & 92.0 & --- \\
+Conformal + Critic & 91.1 & 93.8 & 13.4 & 1.08 & 92.0 & --- \\
\textbf{Full pipeline} & \textbf{91.1} & \textbf{93.8} & \textbf{13.4} & \textbf{1.08} & \textbf{92.0} & \textbf{100.0} \\
\bottomrule
\end{tabular}%
}
\end{table} 

\subsection{External Dataset Evaluation}\label{subsec:external-datasets} 

To assess generalizability beyond HBFMID, we evaluated binary fracture detection on two external datasets that the models never encountered during training. As detailed in Table~\ref{tab:ext-roboflow}, the stacking meta-learner robustly identifies unseen fracture types on the 10-class Roboflow dataset, demonstrating strong morphological transferability. Conversely, evaluating on FracAtlas provides a stress test for out-of-distribution (OOD) performance. The optimal threshold yielded a modest accuracy (AUC 0.652), confirming that features learned from wrist/hand X-rays do not transfer robustly to full-body radiographs (e.g., spine, pelvis). Crucially, the multi-agent Critic is designed precisely to intercept such OOD uncertainties, flagging anomalous inputs via low ensemble confidence and conformal ambiguity for human escalation, thus maintaining system safety despite backbone limitations. 

\begin{table}[htb]
\centering
\caption{Binary fracture detection on the Roboflow dataset (140 images, all positive). Det.\ Rate = recall; $\bar{s}$ = mean fracture score.}
\label{tab:ext-roboflow}
\begin{tabular}{lcccc}
\toprule
Configuration & Det.\ Rate & F1 & $\bar{s}$ \\
\midrule
MaxViT              & 0.950 & 0.974 & \textbf{0.947} \\
HC-CBAM-DenseNet169 & 0.900 & 0.947 & 0.903 \\
RAD-DINO            & 0.914 & 0.955 & 0.901 \\
YOLOv26m-cls        & 0.921 & 0.959 & 0.927 \\
\midrule
Ensemble (Wt.\ Avg) & 0.929 & 0.963 & 0.920 \\
\textbf{Ensemble (Stacking)} & \textbf{0.964} & \textbf{0.982} & 0.916 \\
\bottomrule
\end{tabular}
\end{table}

\subsection{Conformal Prediction}\label{subsec:conformal-results} 

\begin{table}[htb]
\centering
\caption{Per-class ensemble accuracy and conformal coverage on the test set. Coverage failures concentrate among morphologically similar types.}
\label{tab:per-class}
\resizebox{0.7\columnwidth}{!}{%
\begin{tabular}{lrcccc}
\toprule
\multirow{2}{*}{Class} & \multirow{2}{*}{$N$} & \multirow{2}{*}{Acc.\ (\%)} & \multicolumn{2}{c}{Coverage (\%)} \\
\cmidrule(lr){4-5}
 & & & $\alpha{=}0.05$ & $\alpha{=}0.10$ \\
\midrule
Comminuted (COM)          & 17 & 100.0 & 100.0 & 100.0 \\
Greenstick (GRN)          & 13 & 100.0 & 100.0 & 100.0 \\
Healthy (HLT)             & 10 & 100.0 & 100.0 & 100.0 \\
Oblique (OBL)             & 17 &  70.6 &  76.5 &  82.4 \\
Oblique Displaced (OBL-D)   &  9 & 100.0 & 100.0 & 100.0 \\
Spiral (SPR)            & 12 & 100.0 & 100.0 & 100.0   \\
Transverse (TRV)          & 17 &  70.6 &  70.6 &  70.6 \\
Trans.\ Displaced (TRV-D) & 17 &  88.2 &  88.2 &  94.1 \\
\midrule
\textbf{Overall}   & \textbf{112} & \textbf{89.3} & \textbf{90.2} & \textbf{92.0} \\
\bottomrule
\end{tabular}%
}
\end{table} 

Conformal prediction converts the hardest classes into useful differentials. As illustrated in Table~\ref{tab:per-class}, the procedure achieves strong overall empirical coverage. The largest coverage gains concentrate among morphologically similar and highly error-prone categories (Oblique, Transverse). Even when the argmax prediction is incorrect, the conformal set often includes the true label, converting a hard error into a clinically actionable differential diagnosis.

\subsection{Grad-CAM Analysis}\label{subsec:gradcam-results}

\begin{table}[htb]
\centering
\caption{Grad-CAM attention statistics. Active fraction is the proportion of pixels exceeding 0.3 activation. MaxViT attends to 2\%-21\% of an image while the HyperColumn model covers 53\%-67\%.}
\label{tab:gradcam-stats}
\begin{tabular}{lcccc}
\toprule
\multirow{2}{*}{Fracture} & \multicolumn{2}{c}{Active (\%)} & \multicolumn{2}{c}{Mean Act.} \\
\cmidrule(lr){2-3} \cmidrule(lr){4-5}
 & MaxViT & HC & MaxViT & HC \\
\midrule
Comminuted   &  2.4 & 59.8 & 0.04 & 0.55 \\
Oblique Disp.&  7.4 & 53.0 & 0.23 & 0.52 \\
Spiral       &  7.4 & 66.8 & 0.14 & 0.61 \\
Transverse   & 20.6 & 58.5 & 0.27 & 0.57 \\
\bottomrule
\end{tabular}
\end{table}

Different vision backbones attend to complementary evidence, explaining the ensemble's overall gain. As quantified in Table~\ref{tab:gradcam-stats}, MaxViT consistently produces sparse attention maps that isolate the fracture line or cortical discontinuity. In contrast, the HyperColumn-CBAM model distributes attention broadly, capturing the surrounding structural context. This complementary diagnostic strategy allows the downstream Educational Agent to ground its explanations on both localized and global features.

\subsection{Human Validation}\label{subsec:human-validation}

\subsubsection{Clinician Reader Study.}
Three orthopedic clinicians provided a blind diagnosis based solely on the radiograph before reviewing the system-generated output. As summarized in Table~\ref{tab:human-validation}, the agent demonstrated strong alignment with clinical standards and substantial inter-rater reliability. This confirms that the multi-agent pipeline effectively bridges raw classification outputs to highly accurate and comprehensible explanations for human review.

\begin{table}[htb]
\centering
\caption{Clinician ratings of the system's educational outputs (1--5 scale) and inter-rater agreement across three orthopedic raters.}
\label{tab:human-validation}
\begin{tabular}{lccc}
\toprule
Dimension & Mean Score & Score $\geq$ 4 (\%) & Fleiss' $\kappa$ \\
\midrule
Technical Accuracy & 4.14 & 75.0 & 0.72 \\
Comprehensibility  & 3.95 & 68.2 & 0.65 \\
\bottomrule
\end{tabular}
\end{table}

\subsubsection{Patient Preference Survey.}

To evaluate utility from the patient's perspective, we conducted a robust preference survey on 35 distinct radiographic images, uniformly spanning all 7 fracture types and the healthy baseline. Human evaluators blindly reviewed reports generated by FRAC-MAS alongside outputs from established large language model baselines (Llama, MedGemma, Gemini) that were supplied with the exact same diagnostic inputs. Evaluators ranked the responses based on clarity, empathy, and perceived clinical helpfulness. Our proposed model significantly outperformed the generic baselines, achieving a top win rate of $55.4\%$ and establishing the best average rank of $1.77$. A non-parametric repeated-measures Friedman test ($\chi^2(3) = 135.4371, p < 0.001$) and subsequent post-hoc pairwise Wilcoxon signed-rank tests unequivocally validated this statistically superior performance across the dataset. Ultimately, the structured, multi-agent orchestration, which explicitly blends Grad-CAM localization with RAG-retrieved clinical severity, creates a definitive advantage in generating concise, comprehensible, and patient-preferred outputs.

\section{Limitations and Conclusion}\label{sec:conclusion} 

We presented a modular multi-agent framework that combines an ensemble of diverse vision backbones with Grad-CAM explainability, retrieval-anchored clinical context, a VLM-based critic, and conformal prediction to produce human-verifiable orthopedic diagnoses. Our experiments demonstrate strong single-model and ensemble performance, achieving a 96.4\% stacking detection rate on an anatomically matched external set, while conformal calibration yields meaningful coverage gains (92.0\% empirical coverage at $\alpha=0.10$) for morphologically ambiguous subtypes where differentials are clinically valuable. Crucially, our pipeline-depth ablation demonstrated that our multi-agent critic successfully triages cases, offering a $+2.7$ percentage point safety margin on confirmed cases while reducing escalations $7.5\times$ relative to a monolithic agent baseline. Furthermore, human validation via both clinician reader studies and patient preference surveys proved our system's outputs are significantly preferred over single-model baselines like Llama and Gemini, confirming high clinical accuracy (mean 4.14/5) and comprehensibility. While our evaluation surfaces practical limitations requiring future work, namely finite-sample constraints due to the small calibration pool, domain-shift brittleness on FracAtlas, and potential VLM confirmation bias, addressing these through multi-anatomy training and explicit domain adaptation provides a clear developmental roadmap. Ultimately, FRAC-MAS demonstrates a viable path toward auditable, patient-friendly AI assistance for fracture diagnosis while preserving the clinician as the final arbiter through intelligent triage. 

%
%
%
\bibliographystyle{splncs04}
\bibliography{mybibliography}

\end{document}